\documentclass{article}
\usepackage{spconf,amsmath,graphicx,hyperref,spconf,amsmath,graphicx,booktabs,url,hyperref}

\title{Role-guided Speaker Deletion Verification in Clinical Psychiatry Speech Recordings with Audio Language Models}
\makeatletter
\def\name#1{\gdef\@name{#1\\}}
\makeatother

\name{\small\em Joseph T Colonel, Daniel Katzman, Kelsey Kirker, Adam N Davidson, Shalaila S Haas,\\
      \small\em Cheryl Corcoran, Ren\'{e} S Kahn, Guillermo Checci, Baihan Lin%
      }
\address{\small Icahn School of Medicine at Mount Sinai, Department of Psychiatry, New York, NY, USA}

\begin{document}
%\ninept
%
\maketitle

\begin{abstract}
Clinical research in psychiatry increasingly relies on large scale collection of spoken language data to identify acoustic and linguistic biomarkers. Yet evolving consent and protocol requirements can oblige investigators to remove a designated speaker from multi-speaker recordings and to verify said removal at a scale infeasible for manual review of entire corpora. We study this verification problem for role-driven dyadic clinical dialogue in psychiatry and investigate it with two parallel, symmetric pipelines: confirming that clinician speech has been removed from psychiatric interview recordings, and confirming that patient speech has been removed from the same recordings. Each pipeline redacts the raw audio for its target role and then scans the surviving output with audio-language and large-language models to identify missed deletions. We evaluate this approach on a corpus of 48 dyadic recordings drawn from psychiatry settings, testing four open-weight models in an inference-only setting: Gemma-4-12B, Gemma-4-31B, Nemotron-3-Nano, and Nemotron-3-Nano-Omni. A disjunctive OR ensemble over fourteen model-view configurations had a combined F1 of 0.478 (precision 0.330, recall 0.870), an improvement over individual model estimates driven by recall gains that point to substantial complementarity across models and context views.
\end{abstract}

\section{Introduction}
\label{sec:intro}

Large-scale audio corpora have become essential for psychiatric research, enabling identification of speech biomarkers for diagnosis and treatment monitoring~\cite{low2020automated,cummins2015review,haas2026computational,wannan2024accelerating}.%~\cite{low2020automated,cummins2015review,haas2026computational_anonymous,wannan2024accelerating_anonymous}. 
These recordings often capture multiple speakers whose participation may fall under different consent terms and study protocols. Since voiceprint is a HIPAA-protected biometric identifier~\cite{wiepert2024reidentification}, a change in a study's governance requirements or a participant's withdrawal of consent can oblige investigators to remove a designated speaker from multi-speaker recordings at a scale beyond what is feasible for manual review. 

Standard automated deletion pipelines combine diarization, transcription, and role inference to identify and silence the targeted speaker~\cite{okocha2025can}. In general, verifying that deletion succeeded is a separate problem, since typical errors in transcription, diarization, alignment, and speaker-attribution may allow utterances from the intended deleted speaker to survive redaction. In psychiatry this task is made especially challenging due to brief backchanneling common to psychiatric dialogue (e.g. `mmhmm,' `right') that results in short bursts of overlapping speech, affective mirroring in which patient and clinician may adopt similar vocal intonation, as well as patient voice characteristics that shift with symptom severity and emotional state over the course of dialogue~\cite{williamson2013vocal,teferra2022acoustic,france2000acoustical,senaratne2022critical}.

This paper reports an evaluation of open-weight audio-language models for this deletion verification task on a corpus of 48 dyadic clinical recordings drawn from psychiatry. We investigate this task with two parallel, symmetric pipelines applied to the same corpus: clinician deletion, in which clinician audio is redacted using a standard pipeline and the residual is scanned for missed clinician speech; and patient deletion, in which patient audio is redacted and the residual is scanned for missed patient speech. Models are prompted to scan the surviving audio and text transcript for missed deletions.

We evaluate two open-weight audio-language models, spanning audio-capable and text-only configurations, and compare their performance to two large language models in the same model family on both deletion verification tasks. All models are evaluated purely in inference mode, with no training or fine-tuning on this or any other corpus. We show that a disjunctive OR ensemble over  fourteen transcript-guided individual model-view configurations increases combined F1 from 0.408, the best single model configuration, to 0.478, an improvement driven mostly by recall that points to meaningful complementarity in the detections made across models and context views.

\section{Background}
\label{sec:background}

The upstream deletion pipeline evaluated in this work is assembled from standard open-source components (Whisper for transcription, Pyannote 3.1 for diarization, Llama 3.3-70b-instruct for speaker role attribution), and the verification task inherits their characteristic failure modes \cite{okocha2025can,lintoai2023whispertimestamped,radford2022robust,bredin2020pyannote,grattafiori2024llama}. Speaker-assigned  transcriptions are produced by aligning the outputs of Whisper and Pyannote, and Llama 3.3 assigns the roles of clinician and patient to these speakers after analyzing the transcript. From there, audio timeline segments associated with the role targeted for deletion are silenced and the surviving audio is re-transcribed. We summarize relevant lines of work related to these components and position redaction verification with respect to them below.

\subsection{Speaker diarization}
Diarization partitions a recording into speaker-homogeneous segments through voice activity detection, segmentation, speaker embedding extraction, and clustering~\cite{park2022review}. Language-model post-correction of diarization output has also been proposed to repair segment and turn boundaries~\cite{wang2024diarizationlm}. These components determine which turns a deletion pipeline attributes to each speaker, and errors introduced at this stage propagate to the residual leakage this paper detects~\cite{park2022review}. Diarization systems face unique challenges in psychiatry, where affective mirroring, backchanneling, and emotional speech make speaker partitions difficult over long recordings. 

\subsection{Target-speaker extraction and detection}
A related line of work isolates a designated speaker in a recording rather than partitioning all speakers. Target speaker extraction conditions a separation network on an enrollment embedding to recover one voice from a mixture \cite{vzmolikova2019speakerbeam, wang2018voicefilter}. Target-speaker voice activity detection instead labels, at the frame level, when a designated speaker is active~\cite{ding2020personal, medennikov2020target}. These techniques are not considered in this work, as collecting and maintaining a repository of pre-registered voiceprints may be clinically burdensome, inaccurate due to changes in voice relating to symptom severity, or restricted based on consent protocols.

\section{Methods and Data}
\label{sec:methods}

Our corpus comprises 48 dyadic, single-channel clinical recordings drawn from psychiatry and therapy settings, totaling 33 hours and 25 minutes (mean 41 minutes, 46 seconds). Each recording passes through two parallel, symmetric deletion pipelines applied independently to the same raw audio. In the clinician deletion pipeline, the recording is redacted to remove clinician audio, and the surviving output is scanned to identify missed deletions of clinician speech. In the patient deletion pipeline, the same raw recording is separately redacted to remove patient audio, and the surviving output is scanned to identify missed deletions of patient speech. Both pipelines share the same upstream redaction tools and differ only in which role is designated for removal. 

The role-attribution step in both pipelines is grounded in conversational role rather than topic or vocabulary. The verification models evaluated in this paper inherit the same design consideration, and their prompts explicitly instruct the model to attend to acoustic or lexical role cues rather than topic overlap with clinical language, as patients often describe medication and diagnoses in terms that echo their providers.

\subsection{Annotation}
The dyadic corpus was human-annotated for missed deletions after processing through the patient deletion and clinician deletion pipelines. Annotators were instructed not to annotate backchannel events. All annotated leakage events spanned a minimum of one second. Details regarding the full corpus length and total number of annotated events can be found in Table \ref{tab:dataset-size}.

\subsection{Model selection and regulatory scope}
Due to regulatory constraints on this corpus, we were restricted to evaluating open-weight models developed by Google, Microsoft, NVIDIA, and Meta that are run locally on HIPAA-complaint compute. Within this scope, we tested Gemma-4-12B and Gemma-4-31B~\cite{team2024gemma}, Nemotron-3-Nano, and Nemotron-3-Nano-Omni~\cite{deshmukh2026nemotron}. Phi-4-Multimodal~\cite{abdin2024phi} and Audio-Flamingo-Next-Thought~\cite{ghosh2026audio} were also in scope but could not reliably adhere to the prompt instruction's output format during testing and are excluded from analysis. 

\subsection{Model inference and scoring}
\label{subsec:scoring}
Evaluated models were provided different views of the clinician- and patient-deleted transcripts and audio files, three which involved only text, one which combined text with audio, and one of which ingested only audio. For the text-only views, models were provided (1) the full transcript, (2) 300 second-long windows of the transcript with 50\% stride, and (3) 20 second-long windows of the transcript with 50\% stride. For the audio-capable models, a fourth view was included where 20 second-long windows of the timestamped transcript (with 50\% stride) were paired with audio of that segment, and a fifth view where 20 second-long windows of audio (with 50\% stride) were presented without accompanying transcription. Models were specifically tasked with identifying time spans and associated text that contain the missed deletions. Initial testing found that models performed better when prompted to categorize missed deletions according to a fixed set of categories in addition to verbatim flagging. Full prompts can be found in our anonymized repository.\footnote{\href{https://anonymous.4open.science/r/alm_prompts-D031}{https://anonymous.4open.science/r/alm\_prompts-D031}}

After model inference, we merge overlapping flagged intervals before scoring. We then count a true positive for each annotated missed-deletion event overlapped by a merged flagged span, a false negative for each annotated event with no overlap, and a false positive for each merged span with no overlapping annotation. Precision, recall, and F1 are micro-averaged over all sessions.

Following span merging, disjunctive OR ensembles of individual models' flags were evaluated. These ensembles included combining all flagged spans of: all models in all configurations (any); all models scanning only text transcripts (no-audio); audio-capable models in configurations that utilize audio (with-audio); all models excluding the audio-only formulations (transcript-guided); audio-capable models scanning audio and transcript simultaneously (audio-text pairs); audio-capable models scanning only audio (audio-only); all models scanning the entire text transcript in one pass (full-transcript); and all models scanning the text transcript in windows (text-only-windows-300s and text-only-windows-20s). A majority vote of the (any) configuration was additionally considered, where the ensemble only flagged a span as a missed deletion when at least 8 models jointly flagged said span. 

\begin{table}[t]
\centering
\caption{Redaction annotation summary across 48 sessions.}
\label{tab:dataset-size}
\small
\begin{tabular}{lcc}
\toprule
Metric & Clinician & Patient \\
\midrule
Sessions with no missed deletions                  & 9     & 5     \\
Total annotated spans                     & 515   & 872   \\
Spans / session (mean)          & 10.73 & 18.17 \\
Spans / session (median)        & 5.5   & 13.5  \\
Spans / session (range)         & 0--50 & 0--54 \\
Annotated seconds / session (mean)   & 23.4  & 33.1  \\
Annotated seconds / session (median) & 7.5   & 23.0  \\
Total annotated duration (s)    & 1125.0 & 1589.0 \\
\bottomrule
\end{tabular}
\vspace{-.5cm}
\end{table}

\begin{table}[t]
\centering
\caption{Individual model performance, micro-averaged over 48 sessions, broken out by patient deletion, clinician deletion, and combined (pooled) pipelines. For each view, the P/R/F1 triple with the highest F1 is bolded.}
\label{tab:main-results}
\scriptsize
\setlength{\tabcolsep}{2pt}
\resizebox{\columnwidth}{!}{%
\begin{tabular}{l ccc|ccc|ccc}
\toprule
& \multicolumn{3}{c}{\textbf{Patient Del.}} & \multicolumn{3}{c}{\textbf{Clinician Del.}} & \multicolumn{3}{c}{\textbf{Combined}} \\
\cmidrule(lr){2-4}\cmidrule(lr){5-7}\cmidrule(lr){8-10}
Model (mode) & P & R & F1 & P & R & F1 & P & R & F1 \\
\midrule
\midrule
\multicolumn{10}{l}{\textit{Audio only (20s)}} \\
Gemma-4-12B              & 0.057 & 0.302 & 0.096 & 0.038 & 0.153 & 0.061 & 0.051 & 0.247 & 0.085 \\
Nemotron-3-Nano-Omni     & 0.050 & 0.495 & 0.090 & 0.031 & 0.311 & 0.056 & 0.043 & 0.427 & 0.077 \\
\midrule
\multicolumn{10}{l}{\textit{Audio-text pairs (20s)}} \\
Gemma-4-12B              & 0.273 & 0.718 & 0.396 & 0.242 & 0.511 & 0.329 & 0.263 & 0.641 & 0.373 \\
Nemotron-3-Nano-Omni     & 0.291 & 0.739 & 0.418 & 0.223 & 0.542 & 0.316 & 0.266 & 0.665 & 0.381 \\
\midrule
\multicolumn{10}{l}{\textit{Full transcript}} \\
Gemma-4-12B              & 0.291 & 0.609 & 0.394 & 0.202 & 0.210 & 0.206 & 0.271 & 0.461 & 0.341 \\
Gemma-4-31B              & 0.318 & 0.494 & 0.387 & 0.278 & 0.381 & 0.322 & 0.304 & 0.452 & 0.364 \\
Nemotron-3-Nano-Omni     & 0.363 & 0.218 & 0.272 & 0.240 & 0.068 & 0.106 & 0.336 & 0.162 & 0.219 \\
Nemotron-3-Nano          & 0.335 & 0.085 & 0.135 & 0.200 & 0.054 & 0.085 & 0.283 & 0.074 & 0.117 \\
\midrule
\multicolumn{10}{l}{\textit{Windowed transcripts (300s)}} \\
Gemma-4-12B              & \textbf{0.313} & \textbf{0.682} & \textbf{0.429} & 0.268 & 0.474 & 0.342 & 0.298 & 0.605 & 0.400 \\
Gemma-4-31B              & \textbf{0.316} & \textbf{0.667} & \textbf{0.429} & \textbf{0.280} & \textbf{0.540} & \textbf{0.369} & \textbf{0.304} & \textbf{0.620} & \textbf{0.408} \\
Nemotron-3-Nano-Omni     & 0.302 & 0.644 & 0.412 & 0.225 & 0.388 & 0.285 & 0.277 & 0.549 & 0.369 \\
Nemotron-3-Nano          & 0.272 & 0.390 & 0.321 & 0.216 & 0.350 & 0.267 & 0.250 & 0.375 & 0.300 \\
\midrule
\multicolumn{10}{l}{\textit{Windowed transcripts (20s)}} \\
Gemma-4-12B              & 0.295 & 0.760 & 0.425 & 0.246 & 0.612 & 0.351 & 0.277 & 0.705 & 0.398 \\
Gemma-4-31B              & 0.300 & 0.742 & 0.427 & 0.262 & 0.596 & 0.364 & 0.287 & 0.688 & 0.405 \\
Nemotron-3-Nano-Omni     & 0.296 & 0.729 & 0.421 & 0.230 & 0.583 & 0.330 & 0.271 & 0.675 & 0.387 \\
Nemotron-3-Nano          & 0.279 & 0.666 & 0.393 & 0.213 & 0.559 & 0.309 & 0.253 & 0.627 & 0.361 \\
\bottomrule
\vspace{-1.cm}
\end{tabular}%
}
\end{table}

\begin{table}[t]
\centering
\caption{Ensemble performance, micro-averaged over 48 sessions, broken out by patient deletion, clinician deletion, and combined (pooled) pipelines. Ensembles combine the configurations in Table~\ref{tab:main-results} via disjunction (OR) or majority vote. The P/R/F1 triple with the highest F1 is bolded for each view.}
\label{tab:ensemble-results}
\scriptsize
\setlength{\tabcolsep}{2pt}
\resizebox{\columnwidth}{!}{%
\begin{tabular}{l ccc|ccc|ccc}
\toprule
& \multicolumn{3}{c}{\textbf{Patient Del.}} & \multicolumn{3}{c}{\textbf{Clinician Del.}} & \multicolumn{3}{c}{\textbf{Combined}} \\
\cmidrule(lr){2-4}\cmidrule(lr){5-7}\cmidrule(lr){8-10}
Ensemble & P & R & F1 & P & R & F1 & P & R & F1 \\
\midrule
\midrule
OR (any)                     & 0.277 & 0.943 & 0.428 & 0.138 & 0.864 & 0.238 & 0.205 & 0.913 & 0.334 \\
OR (no-audio)                & 0.368 & 0.868 & 0.517 & 0.257 & 0.773 & 0.385 & 0.320 & 0.833 & 0.462 \\
OR (with-audio)              & 0.208 & 0.845 & 0.334 & 0.098 & 0.672 & 0.171 & 0.153 & 0.781 & 0.256 \\
OR (transcript-guided)       & \textbf{0.380} & \textbf{0.898} & \textbf{0.534} & \textbf{0.265} & \textbf{0.823} & \textbf{0.401} & \textbf{0.330} & \textbf{0.870} & \textbf{0.478} \\
OR (audio-text pairs)              & 0.305 & 0.772 & 0.437 & 0.238 & 0.610 & 0.342 & 0.280 & 0.712 & 0.402 \\
OR (audio-only)              & 0.076 & 0.600 & 0.136 & 0.034 & 0.383 & 0.062 & 0.057 & 0.519 & 0.102 \\
OR (full-transcript)         & 0.334 & 0.716 & 0.455 & 0.259 & 0.441 & 0.326 & 0.310 & 0.614 & 0.412 \\
OR (text-only-windows-300s)  & 0.320 & 0.781 & 0.454 & 0.253 & 0.647 & 0.364 & 0.295 & 0.731 & 0.420 \\
OR (text-only-windows-20s)   & 0.315 & 0.833 & 0.457 & 0.246 & 0.748 & 0.371 & 0.287 & 0.801 & 0.423 \\
Majority ($\geq$8/16)        & 0.236 & 0.679 & 0.351 & 0.183 & 0.443 & 0.259 & 0.218 & 0.591 & 0.319 \\
\bottomrule
\vspace{-1cm}
\end{tabular}%
}
\end{table}

\section{Results}
\label{sec:results}

Table \ref{tab:main-results} reports micro-averaged precision, recall, and F1 for all four models across five context views, and Table \ref{tab:ensemble-results} reports the results of ensemble combinations, broken out by patient deletion, clinician deletion, and the pooled, combined pipeline.

Patient deletion was verified more reliably than clinician deletion throughout, though the upstream pipeline properly redacted the clinician more reliably than the patient. Every model and view combination produced a higher F1 for patient deletion than for the matched clinician deletion condition, with gaps in F1 ranging from roughly 0.05 to nearly 0.19 depending on configuration. This pattern held for the ensembles as well. OR (any) reached an F1 of 0.534 on patient deletion against 0.401 on clinician deletion.

Among individual model-view configurations, windowed transcripts outperformed the full transcript view for every model, and the gap was most pronounced for the two Nemotron models. Nemotron-3-Nano fell from a combined F1 of 0.300 on 300-second windows and 0.361 on 20-second windows to 0.117 on the full transcript, a collapse driven almost entirely by recall, which dropped to 0.074 combined. Nemotron-3-Nano-Omni showed the same pattern, falling from combined F1 in the high 0.360s on windowed views to 0.219 on the full transcript. The Gemma models degraded far less on the full transcript view, with Gemma-4-31B reaching a combined F1 of 0.364, close to its windowed performance.

The two audio-text configurations performed comparably to the strongest text-only windowed views but did not exceed them. Nemotron-3-Nano-Omni's audio-text combined F1 of 0.381 sat close to its 20-second windowed transcript F1 of 0.387, suggesting that pairing audio with text at a matched window length offered limited additional benefit over text alone in this setup.

The strongest single, non-ensemble configuration was Gemma-4-31B on 300-second windowed transcripts, with a combined F1 of 0.408 (precision 0.304, recall 0.620). This configuration was also the top performer on both the patient deletion and clinician deletion pipelines individually. Of note, Gemma-4-12B on 20-second windowed transcripts achieved the highest combined recall of all individual model configurations (0.705). 

Ensembling improved substantially on any single configuration.
Disjunctively combining the fourteen model-view configurations that utilized timestamped transcriptions (transcript-guided) performed best, reaching a combined F1 of 0.478 (precision 0.330, recall 0.870). The inclusion of audio-ingesting models marginally improved ensemble F1 in this configuration over the (no-audio) configuration. Ensembling all sixteen model-view configurations yielded the highest combined recall of 0.913.  

\section{Discussion}
\label{sec:discussion}
\subsection{Audio capabilities offer real but limited gains}
The audio-text configurations, evaluated for Gemma-4-12B and Nemotron-3-Nano-Omni at a 20-second window, performed close to but did not clearly exceed the text-only 20-second windowed configuration for either model. This is a modest result given that these models had access to the raw waveform, including prosodic and voice-quality cues that are unavailable to a text-only view. 

Part of the explanation may lie in what these audio-language models were exposed to during instruction tuning. Publicly available audio-instruction corpora rarely include the kind of prosody and affect that characterizes psychiatric speech recordings. A second limitation sits upstream of the models entirely. Because the audio these models ingest has already passed through diarization and role attribution prior to redaction, any residual speech from the redaction target is likely difficult to distinguish from the retained speaker. 

Despite this, audio-ingesting models contributed to the highest performing ensemble (transcript-guided), suggesting that complementary information is captured by analyzing audio alongside transcripts. Strengthening this conclusion, the highest recall ensemble used audio-only configurations. In practice, a high-recall ensemble may be favored as an initial deletion verification tool followed by a round of targeted human review on flagged missed deletions to increase precision.

\subsection{Audio-only models faced a harder task}
Gemma-4-12B and Nemotron-3-Nano-Omni operating in audio-only mode were the two worst individual models evaluated. This may be due  to the more complex task given to them, as these models were not provided a timestamped transcription alongside the audio samples. Instead, these models were simultaneously tasked with internally producing a timestamped transcription and then flagging missed deletions in that transcript. 

We believe these models' severe drop in precision compared to the rest of the individual models is partly due to this challenge. It was found that the audio-only configurations would have timestamp drift between two adjacent analysis windows such that the span-merging described Section \ref{subsec:scoring} would not combine them. Thus though the model may have correctly identified a missed deletion across both windows, the misalignment in timestamps could yield two false positives if neither flagged span aligned with human annotation. 

% \subsection{Clinician deletion verification is harder}
% Every model and view combination in Table \ref{tab:main-results} verified patient deletion more reliably than clinician deletion, and the gap persisted under ensembling. Several features of the corpus plausibly contribute to this asymmetry. Clinician turns are on average shorter and occur more frequently as backchannel-adjacent utterances such as brief acknowledgments and prompts to continue. The annotated corpus also contains fewer clinician spans overall (515 versus 872 for patient, Table \ref{tab:dataset-size}), which may reflect this same tendency for clinician residue to concentrate in short, ambiguous spans.

\subsection{Full-transcript degradation reflects token limits}
The sharp recall collapse for both Nemotron models in the full-transcript view was traced to an output-length limit. Inspection of these models' chain-of-thought outputs in this condition showed that both Nemotron models frequently reached the output token limit before emitting a stop-thinking token and completing the structured JSON response the scoring pipeline requires. The Gemma models were comparably robust to transcript length and did not show this failure mode in their chain-of-thought output. 

\subsection{Regulatory Scope Constrains Model Coverage.}
Our model comparison is limited to open-weight models from four vendors. Whether models built on different audio-encoder architectures or instruction-tuning recipes would change the ranking among modes and ensembles remains open, and a future evaluation with fewer sourcing constraints could address it directly.

\section{Conclusion}
\label{sec:conclusion}

This work demonstrates that audio-language models, in combination with large language models, can be used to verify that a participant has been properly redacted from a psychiatric clinical dialogue recording with high recall. On a 48-session dyadic corpus, a disjunctive OR ensemble over fourteen model-view configurations achieved a combined F1 of 0.478 (precision 0.330, recall 0.870) when identifying where an upstream redaction pipeline missed target participant utterances, an improvement driven by recall gains over individual models that point to substantial complementarity across models and context views. No individual model-view configuration exceeded a combined F1 of 0.408, underscoring that participant deletion verification in psychiatric dyadic recordings benefits from combining text-only and text-audio analysis. Future work should examine whether finetuning audio-capable configurations for this task may improve performance both in verification and upstream pipelines. In addition, widening the audio-text context window toward the 300-second span already available to text-only models is a further direction worth testing directly, since the ambiguous cases these audio-language models must resolve are the same difficult speech segments that the upstream system failed to identify, and longer acoustic context may be what is needed to succeed where the upstream system could not. 
\vfill\pagebreak

\section{Acknowledgments}
We thank the participants and research staff who made this study possible, and colleagues who provided feedback during development. This work is supported by NIH grant 1U01MH136535. Additionally, this work was supported in part through the Minerva computational and data resources and staff expertise provided by Scientific Computing and Data at the Icahn School of Medicine at Mount Sinai and supported by the Clinical and Translational Science Awards (CTSA) grant UL1TR004419 from the National Center for Advancing Translational Sciences. Research reported in this publication was also supported by the Office of Research Infrastructure of the National Institutes of Health under award number S10OD038231. The content is solely the responsibility of the authors and does not necessarily represent the official views of the National Institutes of Health.
{\small
\bibliographystyle{IEEEbib}
\bibliography{strings,refs}
}

\end{document}